\pdfoutput=1
\documentclass{article}
\usepackage{arxiv}
\usepackage[T1]{fontenc}    
\usepackage[colorlinks=true,linkcolor=black,citecolor=black]{hyperref}
\usepackage[utf8]{inputenc}
\usepackage{booktabs}       
\usepackage{amsfonts}       
\usepackage{nicefrac}       
\usepackage{microtype}      

\usepackage[numbers]{natbib}
\usepackage{times,soul}
\usepackage{graphicx}
\usepackage{amsmath,amsfonts,bm,mathtools}
\usepackage{adjustbox,booktabs,colortbl}
\usepackage{url,cleveref}
\usepackage{subcaption}
\usepackage{algorithm,algpseudocode}

\newcommand{\thead}[1]{\multicolumn{1}{c}{\bfseries #1}}
\newcommand{\tcenter}[1]{\multicolumn{1}{c}{#1}}
\newcommand{\tna}{\tcenter{---}}
\graphicspath{{figures/}}
\definecolor{Gray}{gray}{0.9}

\newcommand{\eg}{\textit{e.g\@.}}

\newcommand{\etal}{\textit{et~al\@.}}
\newcommand{\ie}{\textit{i.e\@.}}

\newcommand{\email}[1]{\href{mailto:#1}{#1}}

\newcommand{\cifarx}{CIFAR-10}

\newcommand{\resnetxviii}{ResNet-18}
\newcommand{\resnetl}{ResNet-50}

\newcommand{\mbneti}{MobileNet-V1}
\newcommand{\mbnetii}{MobileNet-V2}

\newcommand{\ordinal}[1]{{#1}^\mathrm{th}}
\newcommand{\tx}{\mathbf{x}}
\newcommand{\ty}{\mathbf{y}}
\newcommand{\weight}{\bm\theta}
\newcommand{\aweight}{\theta}
\newcommand{\cmask}{\mathrm{m}}
\newcommand{\mlehyper}{\bm\phi}
\newcommand{\alphas}{\bm\alpha}
\newcommand{\dataset}{\mathcal{D}}
\newcommand{\loss}{\mathcal{L}}
\newcommand{\losserror}{\loss_\mathrm{E}}
\newcommand{\losscompl}{\loss_\mathrm{C}}
\newcommand{\expect}[2]{\operatorname{\mathbb{E}}_{#1}\left[{#2}\right]}
\newcommand{\kl}[2]{%
    \operatorname{\mathrm{KL}}
    \left({#1} \| {#2}\right)}
\newcommand{\Qshift}[1]{%
    \operatorname{\mathrm{Q}}^\mathrm{shift}_{#1}}
\newcommand{\Qfocus}{%
    \operatorname{\mathrm{Q}}}
\newcommand{\Qrec}{\operatorname{\mathrm{Q}}^{\mathrm{rec}}}
\newcommand{\qshift}[1]{%
    \operatorname{\mathnormal{q}}^\mathrm{shift}_{#1}}
\newcommand{\qfocus}{%
    \operatorname{\mathnormal{q}}_{\mlehyper}}
\newcommand{\qmix}{%
    \operatorname{\mathnormal{q}}_{\mlehyper}^\mathrm{mix}}
\newcommand{\normdist}[2]{%
    \operatorname{\mathcal{N}}\left({#1}, {#2}\right)}
\newcommand{\const}[1]{\mathrm{#1}}
\newcommand{\wasserstein}{\operatorname{\mathcal{W}}}

\title{Focused Quantization for Sparse CNNs}
\author{%
Yiren Zhao\thanks{%
    Xitong Gao and Yiren Zhao
    contributed equally to this work.
    Correspondence to Xitong Gao
    (\email{xt.gao@siat.ac.cn})
    and Yiren Zhao
    (\email{yiren.zhao@cl.cam.ac.uk}).
}\hspace{4pt}\(^1\) \And%
Xitong Gao\footnotemark[1]\hspace{4pt}\(^2\) \And%
Daniel Bates\(^1\) \And%
Robert Mullins\(^1\) \And%
Cheng-Zhong Xu\(^3\) \AND%
\textnormal{\(^1\)~University of Cambridge} \\
\(^2\)~Shenzhen Institutes of Advanced Technology \\
\(^3\)~University of Macau
}

\begin{document}
\maketitle
\begin{abstract}

Deep convolutional neural networks (CNNs)
are powerful tools for a wide range of vision tasks,
but the enormous amount
of memory and compute resources
required by CNNs
pose a challenge in deploying them
on constrained devices.
Existing compression techniques,
while excelling at reducing model sizes,
struggle to be computationally friendly.
In this paper,
we attend to
the statistical properties of sparse CNNs
and present focused quantization,
a novel quantization strategy
based on power-of-two values,
which exploits
the weight distributions after fine-grained pruning.
The proposed method
dynamically discovers
the most effective numerical representation for weights
in layers with varying sparsities,
significantly reducing model sizes.
Multiplications in quantized CNNs
are replaced with much cheaper
bit-shift operations for efficient inference.
Coupled with lossless encoding,
we built a compression pipeline
that provides CNNs with
high compression ratios (CR),
low computation cost
and minimal loss in accuracy.
In \resnetl,
we achieved a \( 18.08 \times \) CR
with only \( 0.24\% \) loss in top-5 accuracy,
outperforming existing compression methods.
We fully compressed a \resnetxviii{}
and found that it is
not only higher in CR and top-5 accuracy,
but also more hardware efficient
as it requires fewer logic gates to implement
when compared to other
state-of-the-art quantization methods
assuming the same throughput.

\end{abstract}

\section{Introduction}\label{sec:intro}

Despite
deep convolutional neural networks (CNNs)
demonstrating state-of-the-art performance
in many computer vision tasks,
their parameter-rich and compute-intensive nature
substantially hinders
the efficient use of them
in bandwidth- and power-constrained devices.
To this end,
recent years have seen
a surge of interest
in minimizing the memory and compute costs
of CNN inference.

\begin{figure}[!ht]
\centering
\newcommand{\sfwidth}{0.33\textwidth}
\begin{subfigure}[b]{\sfwidth}
    \includegraphics[width=\textwidth]{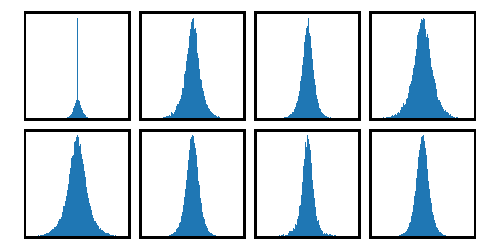}
    \caption{\small Dense layers}\label{fig:intro:dense}
\end{subfigure}%
\begin{subfigure}[b]{\sfwidth}
    \includegraphics[width=\textwidth]{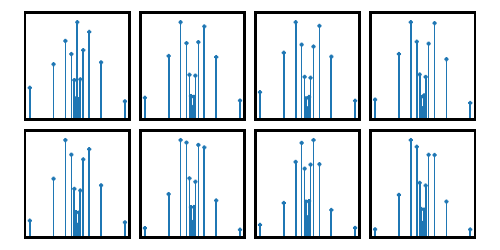}
    \caption{\small After shift quantization}\label{fig:intro:dense_quantized}
\end{subfigure} \\
\begin{subfigure}[b]{\sfwidth}
    \includegraphics[width=\textwidth]{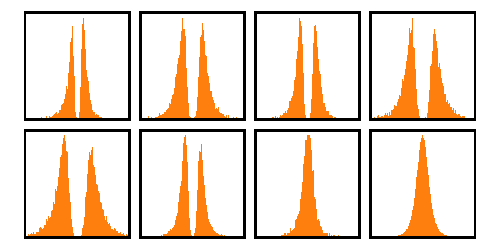}
    \caption{\small Sparse layers }\label{fig:intro:sparse}
\end{subfigure}%
\begin{subfigure}[b]{\sfwidth}
    \includegraphics[width=\textwidth]{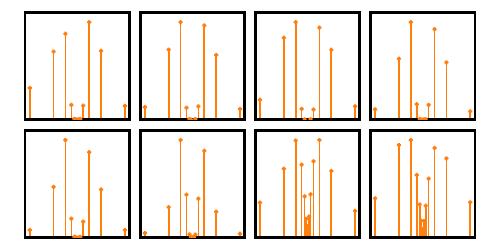}
    \caption{\small After shift quantization}\label{fig:intro:sparse_quantized}
\end{subfigure}
\caption{%
    The weight distributions
    of the first 8 layers of \resnetxviii{} on ImageNet.
    (\protect\subref{fig:intro:dense})
    shows the weight distributions of the layers,
    (\protect\subref{fig:intro:sparse})
    similarly shows the distributions (excluding zeros)
    for a sparsified variant.
    (\protect\subref{fig:intro:dense_quantized}) and
    (\protect\subref{fig:intro:sparse_quantized})
    respectively quantize the weight distributions
    on the left with 5-bit shift quantization.
    Note that in some sparse layers,
    greedy pruning encourages
    weights to avoid near zero values.
    Shift quantization on these layers
    thus results in poor utilization
    of the quantization levels.
}\label{fig:intro}
\end{figure}

Pruning algorithms compress CNNs
by setting weights to zero,
thus removing connections or neurons
from the models.
In particular,
fine-grained pruning~\citep{liu2015sparse,guo2016dynamic}
provides the best compression
by removing connections
at the finest granularity,
\ie~individual weights.
Quantization methods
reduce the number of bits
required to represent each value,
and thus further provide
memory, bandwidth and compute savings.
\emph{Shift quantization} of weights,
which quantizes weight values in a model
to powers-of-two or zero,
\ie~\( \{ 0, \pm1, \pm2, \pm4, \ldots \} \),
is of particular of interest,
as multiplications in convolutions
become much-simpler
bit-shift operations.
The computational cost in hardware
can thus be significantly reduced
without a detrimental impact
on the model's task accuracy~\citep{zhou2017incremental}.
Fine-grained pruning, however,
is often in conflict with quantization,
as pruning introduces
various degrees of sparsities
to different layers~\citep{zhao2018mayo,nikolic2019}.
Linear quantization methods
(integers) have uniform quantization levels
and non-linear quantizations
(logarithmic, floating-point and shift)
have fine levels around zero but levels
grow further apart as values get larger in magnitude.
Both linear and nonlinear quantizations thus
provide precision where
it is not actually required
in the case of a pruned CNN\@.
It is observed that empirically,
very few non-zero weights
concentrate around zero
in some layers that are sparsified
with fine-grained pruning
(see \Cref{fig:intro:sparse} for an example).
Shift quantization is highly desirable
as it can be implemented efficiently,
but it becomes a poor choice
for certain layers in sparse models,
as most near-zero quantization levels
are under-utilized (\Cref{fig:intro:sparse_quantized}).

\begin{figure}[!ht]
\centering
\includegraphics[width=1.0\textwidth]{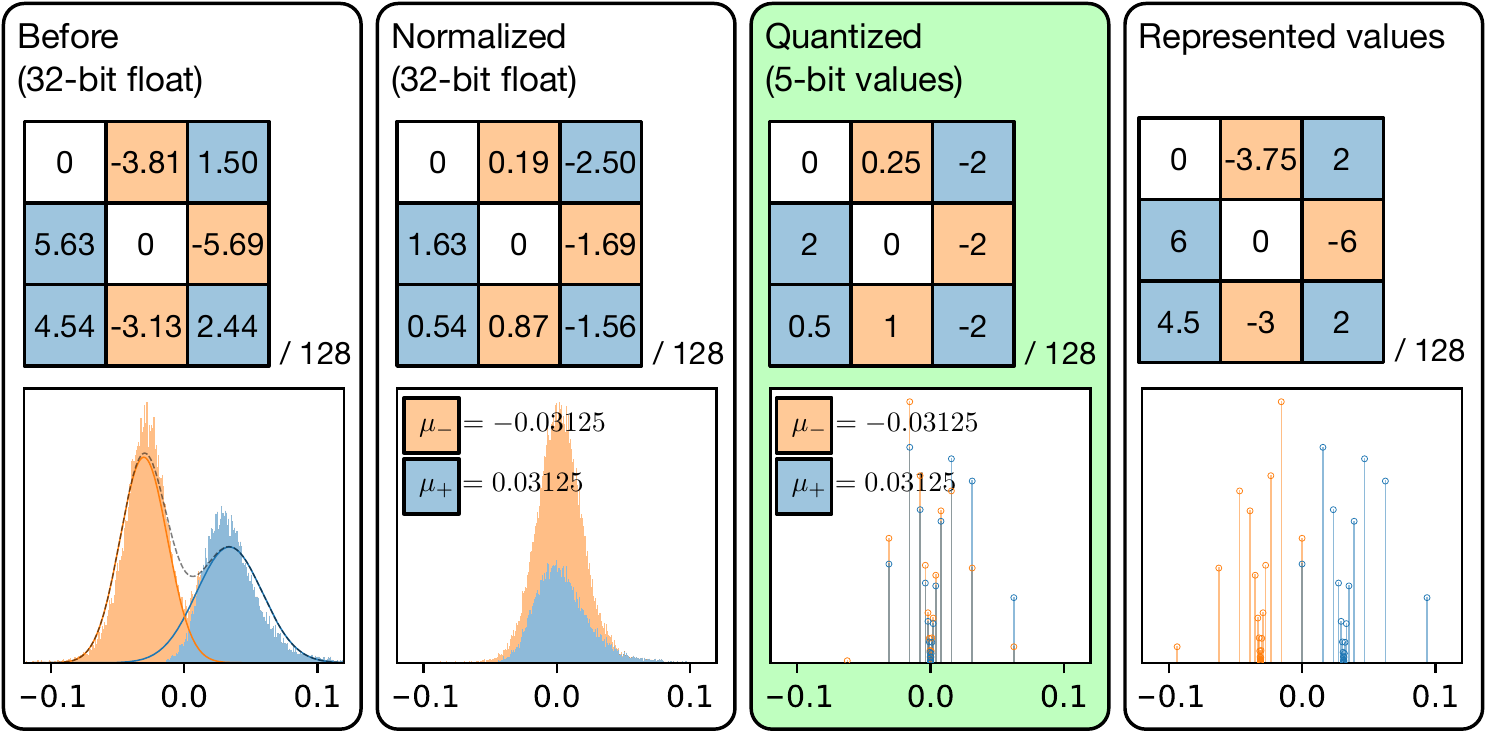}
\caption{%
    The step-by-step process
    of recentralized quantization
    of unpruned weights
    on \texttt{block3f/conv1} in sparse \resnetl{}.
    Each step shows how it changes a filter
    and the distribution of weights.
    Higher peaks in the histograms
    denote values found with higher frequency.
    Values in the filter
    share a common denominator 128,
    indicated by ``\( / 128 \)''.
    The first estimates
    the high-probability regions
    with a Gaussian mixture,
    and assign weights to a Gaussian component.
    The second normalizes each weight.
    The third quantizes
    the normalized values with shift quantization
    and produces a representation of quantized weights
    used for inference.
    The final block visualizes
    the actual numerical values after quantization.
}\label{fig:mle:overview}
\end{figure}

This dichotomy prompts the question,
\emph{
    how can we quantize sparse weights
    efficiently and effectively?
}
Here, efficiency represents
not only the reduced model size
but also the minimized compute cost.
Effectiveness means that
the quantization levels are well-utilized.
From an information theory perspective,
it is desirable to
design a quantization function \( Q \)
such that the quantized values in
\( \hat{\weight} = Q(\weight) \)
closely match the prior weight distribution.
We address both issues
by proposing a new approach
to quantize parameters in CNNs
which we call \emph{focused quantization} (FQ)
that mixes \emph{shift}
and \emph{recentralized} quantization methods.
Recentralized quantization
uses a mixture of Gaussian distributions
to find the most concentrated probability masses
in the weight distribution of sparse layers
(first block in \Cref{fig:mle:overview}),
and independently quantizes the probability masses
(rightmost of \Cref{fig:mle:overview})
to powers-of-2 values.
Additionally,
not all layers consist of
two probability masses,
and recentralized quantization may not be necessary
(as shown in \Cref{fig:intro:sparse}).
In such cases,
we use the Wasserstein distance
between the two Gaussian components
to decide when to apply shift quantization.

For evaluation, we present
a complete compression pipeline
comprising fine-grain pruning,
FQ and Huffman encoding
and estimate the resource utilization
in custom hardware required for inference.
We show that
the compressed models with FQ
not only provide higher task accuracies,
but also require less storage
and lower logic usage
when compared to other methods.
This suggests the FQ-based compression
is a more practical alternative design
for future custom hardware accelerators
designed for neural network inference~\citep{zhao2019automatic}.

In this paper, we make the following contributions:
\begin{itemize}

    \item
    The proposed method,
    focused quantization for sparse CNNs,
    significantly reduces both computation and model size
    with minimal loss of accuracy.

    \item FQ is hybrid,
    it systematically mixes
    a recentralized quantization
    with shift quantization
    to provide the most effective quantization
    on sparse CNNs.

    \item We built
    a complete compression pipeline based on FQ\@.
    We observed that FQ achieves
    the highest compression rates
    on a range of modern CNNs
    with the least accuracy losses.

    \item We found that
    a hardware design based on FQ
    demonstrates the most efficient
    hardware utilization compared to
    previous state-of-the-art
    quantization methods~\citep{lin2017towards,zhanglq2018}.

\end{itemize}
The rest of the paper
is structured as follows.
\Cref{sec:related}
discusses related work
in the field of model compression.
\Cref{sec:method}
introduces focused quantization.
\Cref{sec:eval}
presents and evaluates the proposed compression pipeline
and \Cref{sec:conclusion} concludes the paper.

\section{Related Work}\label{sec:related}

Recently, a wide range of techniques
have been proposed and proven effective
for reducing the memory
and computation requirements of CNNs.
These proposed optimizations
can provide direct reductions
in memory footprints, bandwidth requirements,
total number of arithmetic operations,
arithmetic complexities
or a combination of these properties.

Pruning-based optimization methods directly
reduce the number of parameters in a network.
Fine-grained pruning method~\citep{guo2016dynamic}
significantly reduces the size
of a model but introduces element-wise sparsity.
Coarse-grained pruning~\citep{hao2017thinet,gao2018dynamic}
shrinks model sizes and reduce computation
at a higher granularity
that is easier to accelerate
on commodity hardware.
Quantization methods
allow parameters to be represented
in more efficient data formats.
Quantizing weights
to powers-of-2 recently gained attention because
it not only reduces the model size
but also simplifies computation
\citep{leng2018extremely,zhou2017incremental,%
miyashitaLM16,zhao2019automatic}.
Previous research also focused
on quantizing CNNs to extremely low bit-widths
such as ternary~\citep{zhu2016trained}
or binary~\citep{hubara2016binarized} values.
They however introduce large numerical errors
and thus cause significant degradations
in model accuracies.
To minimize loss in accuracies,
the proposed methods
of \cite{zhanglq2018} and \cite{lin2017towards}
quantize weights to \( N \) binary values,
compute \( N \) binary convolutions
and scale the convolution outputs
individually before summation.
Lossy and lossless encodings
are other popular methods
to reduce the size of a DNN,
typically used in conjunction
with pruning and quantization~\citep{dubey2018coreset,han2015deep}.

Since many compression techniques are available
and building a compression pipeline
provides a multiplying effect
in compression ratios,
researchers start to chain
multiple compression techniques.
Han~\etal~\cite{han2015deep} proposed
Deep Compression that combines
pruning, quantization and Huffman encoding.
Dubey~\etal~\cite{dubey2018coreset} built
a compression pipeline using
their coreset-based filter representations.
Tung~\etal~\cite{tung2018clipq}
and Polino~\etal~\cite{polino2018model}
integrated multiple compression techniques,
where \cite{tung2018clipq}
combined pruning with quantization
and \cite{polino2018model}
employed knowledge distillation on top of quantization.
Although there are many attempts
in building an efficient compression pipeline,
the statistical relationship
between pruning and quantization lacked exploration.
In this paper,
we look at exactly this problem
and propose a new method that exploits
the statistical properties
of weights in pruned models
to quantize them efficiently and effectively.

\section{Method}\label{sec:method}

\subsection{%
    Preliminaries: Shift quantization
}\label{sec:method:preliminaries}


Shift quantization is a quantization scheme
which constrains weight values
to powers-of-two or zero values.
A representable value in a \( (\const{k} + 2) \)-bit
shift quantization is given by:
\begin{equation}
    \label{eq:shift}
    v = s \cdot 2^{e - \const{b}},
\end{equation}
where \( s = \{ -1, 0, 1 \} \)
denotes either zero or the sign of the value,
\( e \) is an integer
bounded by \( [0, 2^\const{k} - 1] \),
and \( \const{b} \) is the bias,
a layer-wise constant which scales
the magnitudes of quantized values.
We use \( \hat{\aweight} = \Qshift{n, \const{b}}[\aweight] \)
to denote a \( \const{n} \)-bit shift quantization
with a bias \( \const{b} \)
of a weight value \( \aweight \)
to the nearest representable value \( \hat{\aweight} \).
As we have discussed earlier
and illustrated in \Cref{fig:intro},
shift quantization on sparse layers
makes poor use of the range of representable values,
\ie~the resulting distribution after quantization
\( \qshift{\const{n}, \const{b}} (\aweight) \)
is a poor approximation
of the original layer weight distribution
\( p(\aweight | \dataset) \),
where \( \dataset \) is the training dataset.

\subsection{Designing the Recentralized Quantization Function}

Intuitively, it is desirable
to concentrate quantization effort
on the high probability regions
in the weight distribution in sparse layers.
By doing so,
we can closely match the distribution
of quantized weights with the original,
and thus at the same time
incur smaller round-off errors.
Recentralized quantization \( \Qfocus[\aweight] \)
is designed specifically for this purpose,
and applied in a layer-wise fashion.
Assuming that \( \aweight \in \weight \) is
a weight value of a convolutional layer,
we can define \( \Qfocus[\aweight] \) as follows:
\begin{equation}
    \label{eq:focused_quantization}
    \Qfocus[\aweight] =
        z_\theta \alpha \sum_{c \in C}
            \delta_{c, \cmask_{\aweight}}
            \Qrec_c[\aweight],
    \text{~where~}
    \Qrec_c[\aweight] =
        \Qshift{\const{n}, \const{b}}\left[
            \frac{\aweight - \mu_c}{\sigma_c}
        \right] \sigma_c + \mu_c.
\end{equation}
Here \( z_\aweight \) is a predetermined constant
\( \{ 0, 1 \} \) binary value
to indicate if \( \aweight \) is pruned,
and it is used to set pruned weights to 0.
The set of components \( c \in C \)
determines the locations
to focus quantization effort,
each specified by the component's
mean \( \mu_c \) and standard deviation \( \sigma_c \).
The Kronecker delta
\( \delta_{c, \cmask_\aweight} \)
evaluates to either 1
when \( c = \cmask_\aweight \),
or 0 otherwise.
In other words,
the constant \( \cmask_\aweight \in C \)
chooses which component in
\( C \) is used to quantize \( \aweight \).
Finally, \( \Qrec_c[\aweight] \)
locally quantizes the component \( c \)
with shift quantization.
Following~\cite{zhu2016trained}
and \cite{leng2018extremely},
we additionally introduce
a layer-wise learnable scaling factor \( \alpha \)
initialized to 1,
which empirically improves the task accuracy.

By adjusting the \( \mu_c \) and \( \sigma_c \)
of each component \( c \),
and finding suitable assignments
of weights to the components,
the quantized weight distribution \( \qfocus(\aweight) \)
can thus match the original closely,
where we use \( \mlehyper \)
as a shorthand to denote the relevant hyperparameters,
\eg~\( \mu_c \), \( \sigma_c \).
The following section explains
how we can optimize them efficiently.

\subsection{%
    Optimizing Recentralized Quantization
    \texorpdfstring{\( \Qfocus[\aweight] \)}{}
}\label{sec:method:optimizing_focused}

Hyperparameters \( \mu_c \) and \( \sigma_c \)
in recentralized quantization
can be optimized by applying the following
two-step process in a layer-wise manner,
which first identifies regions
with high probabilities
(first block in \Cref{fig:mle:overview}),
then locally quantize them
with shift quantization
(second and third blocks in \Cref{fig:mle:overview}).
First, we notice that in general,
the weight distribution resembles
a mixture of Gaussian distributions.
It is thus more efficient to
find a Gaussian mixture model \( \qmix(\aweight) \)
that approximates the original distribution
\( p(\aweight | \dataset) \)
to closely optimize the above objective:
\begin{equation}
    \label{eq:surrogate}
    \qmix(\aweight) = \sum_{c \in C}
        \lambda_c f(\aweight | \mu_c, \sigma_c),
\end{equation}
where \( f(\aweight | \mu_c, \sigma_c) \)
is the probability density function
of the Gaussian distribution
\( \normdist{\mu_c}{\sigma_c} \),
the non-negative \( \lambda_c \)
defines the mixing weight
of the \( \ordinal{c} \) component
and \( \operatorname\Sigma_{c \in C} \lambda_c \) \( = 1 \).
Here, we find the set of hyperparameters
\( \mu_c \), \( \sigma_c \) and \( \lambda_c \)
contained in \( \mlehyper \)
that maximizes \( \qmix(\aweight) \)
given that \( \aweight \sim p(\aweight | \dataset) \).
This is known as the \emph{maximum likelihood estimate} (MLE),
and the MLE can be efficiently computed
by the \emph{expectation-maximization} (EM)
algorithm~\citep{dempster77mle}.
In practice,
we found it sufficient to
use two Gaussian components,
\( C = \{-, +\} \),
for identifying high-probability regions
in the weight distribution.
For faster EM convergence,
we initialize \( \mu_-, \sigma_- \)
and \( \mu_+, \sigma_+ \) respectively
with the means and standard deviations
of negative and positive values
in the layer weights respectively,
and \( \lambda_-, \lambda_+ \) with \( \frac12 \).

We then generate \( \cmask_\aweight \)
from the mixture model,
which individually selects
the component to use for each weight.
For this,
\( \cmask_\aweight \) is evaluated
for each \( \aweight \)
by sampling a categorical distribution
where the probability
of assigning a component \( c \)
to \( \cmask_\aweight \),
\ie~\( p( \cmask_\aweight = c )\),
is \(
    \lambda_c f(\aweight | \mu_c, \sigma_c)
    / \qmix(\aweight) \).

Finally, we set the constant \( \const{b} \)
to a powers-of-two value,
chosen to ensure that \(
    \qshift{\const{n}, \const{b}} \left[ \cdot \right]
\) allows at most a proportion of
\( \frac{1}{2^\const{n} + 1} \) values to overflow
and clips them to the maximum representable magnitude.
In practice,
this heuristic choice makes better use
of the quantization levels
provided by shift quantization
than disallowing overflows.
After determining
all of the relevant hyperparameters
with the method described above,
\( \hat{\aweight} = \Qfocus[\aweight] \)
can be evaluated
to quantize the layer weights.

\subsection{Choosing the Appropriate Quantization}\label{sec:method:wsep}

\begin{figure}[ht]
\centering
\begin{subfigure}[c]{0.3\linewidth}
    \centering
    \includegraphics[scale=0.6]{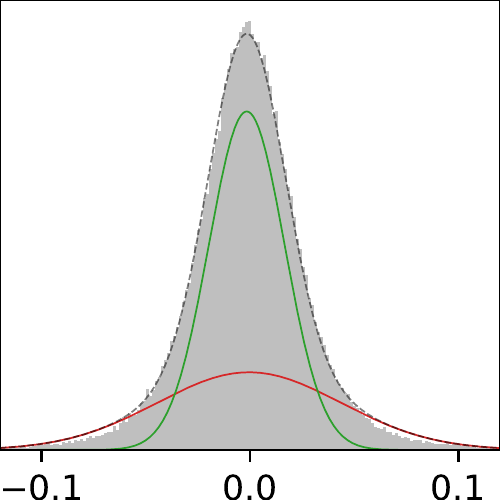}
    \caption{%
        \small Weight distribution.
    }\label{fig:unseparated:before}
\end{subfigure}
\begin{subfigure}[c]{0.3\linewidth}
    \centering
    \includegraphics[scale=0.6]{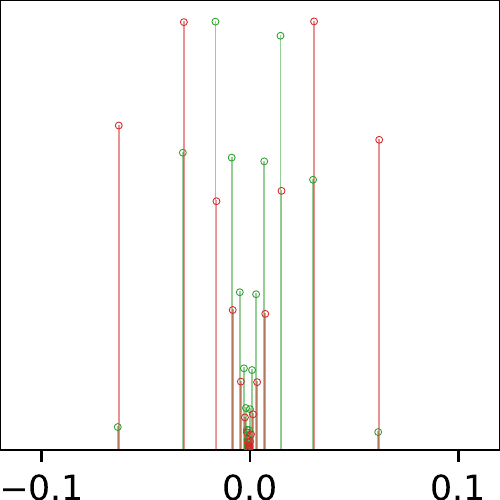}
    \caption{%
        \small Overlapping components.
    }\label{fig:unseparated:after}
\end{subfigure}
\caption{%
    The weight distribution of
    the layer \texttt{block22/conv1}
    in a sparse \resnetxviii{} trained on ImageNet,
    as shown by the histograms.
    It shows that
    when the two Gaussian components
    have a large overlap,
    quantizing with either one of them
    results in almost the same quantization levels.}
\end{figure}

As we have discussed earlier,
the weight distribution of sparse layers
may not always have multiple high-probability regions.
For example,
fitting a mixture model of two Gaussian components
on the layer in \Cref{fig:unseparated:before}
gives highly overlapped components.
It is therefore of little consequence
which component we use to
quantize a particular weight value.
Under this scenario,
we can simply use \( \const{n} \)-bit shift quantization
\( \Qshift{\mathrm{n}, \mathrm{b}}[\cdot] \)
instead of a \( \const{n} \)-bit \( \Qfocus[\cdot] \)
which internally uses
a \( (\const{n} - 1) \)-bit signed shift quantization.
By moving the 1 bit
used to represent the now absent \( \cmask \)
to shift quantization,
we further increase its precision.

To decide whether to use
shift or recentralized quantization,
it is necessary to introduce a metric
to compare the similarity
between the pair of components.
While the KL-divergence
provides a measure for similarity,
it is however non-symmetric,
making it unsuitable for this purpose.
To address this,
we propose to
first normalize the distribution of the mixture,
then to use the 2-Wasserstein metric
between the two Gaussian components after normalization
as a decision criterion,
which we call the \emph{Wasserstein separation}:
\begin{equation}
    \label{eq:wasserstein}
    \wasserstein(c_1, c_2) = \frac{1}{\sigma^2} \left(
        {(\mu_{c_1} - \mu_{c_2})}^2 +
        {(\sigma_{c_1} - \sigma_{c_2})}^2
    \right),
\end{equation}
where \( \mu_c \) and \( \sigma_c \)
are respectively the mean and standard deviation
of the component \( c \in \{ c_1, c_2 \} \),
and \( \sigma^2 \) denotes the variance
of the entire weight distribution.
FQ can then adaptively pick to
use recentralized quantization
for all sparse layers
except when \( \wasserstein(c_1, c_2) < w_\mathrm{sep} \),
and shift quantization is used instead.
In our experiments,
we found \( w_\mathrm{sep} = 2.0 \)
usually provides a good decision criterion.
In \Cref{sec:eval:wsep},
we additionally study
the impact of quantizing a model
with different \( w_\mathrm{sep} \) values.

\subsection{Model Optimization}\label{sec:method:optimization}

To optimize the quantized sparse model,
we integrate the quantization process described above
into the gradient-based training of model parameters.
Initially, we compute the hyperparameters
\( \mu_c, \sigma_c, \lambda_c \) for each layer,
and generate the component selection mask
\( \cmask_\aweight \) for each weight \( \aweight \)
with the method in \Cref{sec:method:optimizing_focused}.
The resulting model is then fine-tuned
where the forward pass uses quantized weights
\( \hat\aweight = \Qfocus[\aweight] \),
and the backward pass updates
the floating-point weight parameters \( \aweight \)
by treating the quantization as an identity function.
During the fine-tuning process,
the hyperparameters used by \( \Qfocus[\aweight] \)
are updated using the current weight distribution
at every \( k \) epochs.
We also found that in our experiments,
exponentially increasing the interval \( k \)
between consecutive hyperparameter updates
helps to reduce the variance introduced by sampling
and improves training quality.

\subsection{The MDL Perspective}

Theoretically, the model optimization
can be formulated
as a \emph{minimum description length} (MDL)
optimization~\citep{hinton93mdl,graves11}.
Given that we approximate
the posterior \( p(\aweight | \dataset) \)
with a distribution of quantized weights
\( \qfocus(\aweight) \),
where \( \mlehyper \) contains
the hyperparameters used
by the quantization function \( \Qfocus[\aweight] \),
the MDL problem minimizes
the \emph{variational free energy}~\citep{graves11},
\(
    \loss(\weight, \alphas, \mlehyper)
    = \losserror + \losscompl
\), where:
\begin{equation}
    \label{eq:free_energy}
    \losserror = \expect{\hat\weight \sim \qfocus(\aweight)}{
        -\log p(\ty | \tx, \alphas, \hat\weight)
    }, \quad
    \losscompl = \kl{\qfocus(\aweight)}{p(\aweight | \dataset)}.
\end{equation}
The error cost \( \losserror \)
reflects the cross-entropy loss
of the quantized model,
with quantized weights \( \hat\weight \)
and layer-wise scalings \( \alphas \),
trained on the dataset \( \dataset = ( \tx, \ty ) \),
which is optimized by stochastic gradient descent.
The complexity cost \( \losscompl \)
is the \emph{Kullback-Leibler} (KL) divergence
from the quantized weight distribution
to the original.
Intuitively, minimizing \( \losscompl \)
reduces the discrepancies
between the weight distributions
before and after quantization.
As this is intractable,
we replace \( \qfocus(\aweight) \)
with a close surrogate,
a Gaussian mixture \( \qmix(\aweight) \).
It turns out that the process of finding the MLE
discussed in \Cref{sec:method:optimizing_focused}
is equivalent to minimizing
\( \displaystyle \kl{\qmix(\aweight)}{p(\aweight | \dataset)} \),
a close proxy for \( \losscompl \).
\Cref{sec:method:optimization} then interleaves
the optimization of \( \losserror \) and \( \losscompl \)
to minimize the MDL objective
\( \loss(\weight, \alphas, \mlehyper) \).

\section{Evaluation}\label{sec:eval}


We applied \emph{focused compression} (FC),
a compression flow which consists of
pruning, FQ and Huffman encoding,
on a wide range of popular vision models including
MobileNets~\citep{howard2017mobilenets,sandler2018mobilenetv2}
and ResNets~\citep{he2016resnet,he2016identity}
on the ImageNet dataset~\citep{deng2009imagenet}.
For all of these models,
FC produced models
with high compression ratios (CRs)
and permitted a multiplication-free
hardware implementation of convolution
while having minimal impact on the task accuracy.
In our experiments,
models are initially sparsified
using Dynamic Network Surgery~\citep{guo2016dynamic}.
FQ is subsequently applied
to restrict weights to low-precision values.
During fine-tuning, we additionally employed
Incremental Network Quantization
(INQ)~\citep{zhou2017incremental}
and gradually increased
the proportion of weights being quantized
to 25\%, 50\%, 75\%, 87.5\% and 100\%.
At each step,
the models were fine-tuned for 3 epochs
at a learning rate of 0.001,
except for the final step at 100\%
we ran for 10 epochs,
and decay the learning rate every 3 epochs.
Finally, Huffman encoding
was applied to model weights
which further reduced model sizes.
To simplify inference computation
in custom hardware (\Cref{sec:eval:compute}),
in our experiments
\( \mu_- \) and \( \mu_+ \)
are quantized to the nearest powers-of-two values,
and \( \sigma_- \) and \( \sigma_+ \)
are constrained to be equal.

\subsection{Model Size Reduction}\label{sec:eval:size}

\Cref{tab:results}
compares the accuracies and compression rates
before and after applying the compression pipeline
under different quantization bit-widths.
It demonstrates the effectiveness
of FC on the models.
We found that sparsified ResNets with 7-bit weights
are at least \( 16\times \) smaller
than the original dense model
with marginal degradations
(\( \leq \hspace{-0.3em} 0.24\% \))
in top-5 accuracies.
MobileNets,
which are much less redundant
and more compute-efficient models to begin with,
achieved a smaller CR at around \( 8\times \)
and slightly larger accuracy degradations
(\( \leq \hspace{-0.3em} 0.89\% \)).
Yet when compared to the \resnetxviii{} models,
it is not only more accurate,
but also has a significantly
smaller memory footprint at 1.71 MB\@.

In \Cref{tab:comp} we compare FC
with many state-of-the-art model compression schemes.
It shows that FC
simultaneously achieves the best accuracies
and the highest CR on both ResNets.
Trained Ternary Quantization (TTQ)~\citep{zhu2016trained}
quantizes weights to ternary values,
while INQ~\citep{zhou2017incremental}
and extremely low bit neural network
(denoted as ADMM)~\citep{leng2018extremely}
quantize weights to
ternary or powers-of-two values
using shift quantization.
Distillation and Quantization (D\&Q)~\citep{polino2018model}
quantize parameters to integers via distillation.
Note that D\&Q's results
used a larger model as baseline,
hence the compressed model
has high accuracies and low CR\@.
We also compared against
Coreset-Based Compression~\citep{dubey2018coreset}
comprising pruning, filter approximation,
quantization and Huffman encoding.
For \resnetl,
we additionally compare against ThiNet~\citep{hao2017thinet},
a filter pruning method,
and Clip-Q~\citep{tung2018clipq},
which interleaves training steps
with pruning, weight sharing and quantization.
FC again achieves
the highest CR (\( 18.08\times \)) and accuracy (74.86\%).

\begin{table}[!ht]
\centering
\caption{%
    The accuracies (\%), sparsities (\%) and CRs
    of focused compression on ImageNet models.
    The baseline models are
    dense models before compression
    and use 32-bit floating-point weights,
    and 5 bits and 7 bits
    denote the number of bits used by individual weights
    of the quantized models before Huffman encoding.
}\label{tab:results}
\adjustbox{scale=1.0}{%
\begin{tabular}{lrlrlrrrr}
\toprule
\thead{Model}
    & \thead{Top-1} & \thead{\( \Delta \)}
    & \thead{Top-5} & \thead{\( \Delta \)}
    & \thead{Sparsity} & \thead{Size (MB)} & \thead{CR (\( \times \))} \\
\midrule
\resnetxviii{} & 68.94 & \tna{} & 88.67 & \tna{} &  0.00 & 46.76 & \tna{} \\
Pruned         & 69.24 & \hphantom{-}0.30  & 89.05 & \hphantom{-}0.38  & 74.86 & 8.31  &  5.69  \\
5 bits         & 68.36 & -0.58  & 88.45 & -0.22  & 74.86 & 2.86  & 16.33  \\
7 bits         & 68.57 & -0.37  & 88.53 & -0.14  & 74.86 & 2.94  & 15.92  \\
\midrule
\resnetl{}     & 75.58 & \tna{} & 92.83 & \tna{} &  0.00 & 93.82 & \tna{} \\
Pruned         & 75.10 & -0.48  & 92.58 & -0.25  & 82.70 & 11.76 &  7.98  \\
5 bits         & 74.86 & -0.72  & 92.59 & -0.24  & 82.70 & 5.19  & 18.08  \\
7 bits         & 74.99 & -0.59  & 92.59 & -0.24  & 82.70 & 5.22  & 17.98  \\
\midrule
\mbneti{}      & 70.77 & \tna{} & 89.48 & \tna{} &  0.00 & 16.84 & \tna{} \\
Pruned         & 70.03 & -0.74  & 89.13 & -0.35  & 33.80 & 6.89  & 2.44   \\
7 bits         & 69.13 & -1.64  & 88.61 & -0.87  & 33.80 & 2.13  & 7.90   \\
\midrule
\mbnetii{}     & 71.65 & \tna{} & 90.44 & \tna{} &  0.00 & 13.88 & \tna{} \\
Pruned         & 71.24 & -0.41  & 90.31 & -0.13  & 31.74 & 5.64  & 2.46   \\
7 bits         & 70.05 & -1.60  & 89.55 & -0.89  & 31.74 & 1.71  & 8.14   \\
\bottomrule
\end{tabular}%
}
\end{table}

\begin{table}[!ht]
\newcommand{\asterisk}{\( ^\star \)\hspace{-4.5pt}}
\centering
\caption{%
    Comparisons of top-1 and top-5 accuracies (\( \% \)) and CRs
    with various compression methods.
    Numbers with \( ^\star \) indicate
    results not originally reported and calculated by us.
    Note that D\&Q used a much larger \resnetxviii{},
    the 5 bases used by ABC-Net
    denote 5 separate binary convolutions.
    LQ-Net used a ``pre-activation''
    \resnetxviii{}~\citep{he2016identity}
    with a 1.4\% higher accuracy baseline than ours.
}\label{tab:comp}
\adjustbox{scale=1.0}{%
\begin{tabular}{lrrrr}
\toprule
\thead{\resnetxviii}
    & \thead{Top-1} & \thead{Top-5}
    & \thead{Size (MB)} & \thead{CR (\( \times \))} \\
\midrule
TTQ~\citep{zhu2016trained}
    & 66.00
    & 87.10
    & 2.92\asterisk{}
    & 16.00\asterisk{} \\
INQ (2 bits)~\citep{zhou2017incremental}
    & 66.60
    & 87.20
    & 2.92\asterisk{}
    & 16.00\asterisk{} \\
INQ (3 bits)~\citep{zhou2017incremental}
    & 68.08
    & 88.36
    & 4.38\asterisk{}
    & 10.67\asterisk{} \\
ADMM (2 bits)~\citep{leng2018extremely}
    & 67.0\hphantom{0}
    & 87.5\hphantom{0}
    & 2.92\asterisk{}
    & 16.00\asterisk{} \\
ADMM (3 bits)~\citep{leng2018extremely}
    & 68.0\hphantom{0}
    & 88.3\hphantom{0}
    & 4.38\asterisk{}
    & 10.67\asterisk{} \\
ABC-Net (5 bases, or 5 bits) \citep{lin2017towards}
    & 67.30
    & 87.90
    & 7.30\asterisk{}
    & 6.4\hphantom{0}\asterisk{} \\
LQ-Net (preact, 2 bits) \citep{zhanglq2018}
    & 68.00
    & 88.00
    & 2.92\asterisk{}
    & 16.00\asterisk{} \\
D\&Q (large)~\citep{polino2018model}
    & \textbf{73.10}
    & \textbf{91.17}
    & 21.98
    & 2.13\asterisk{} \\
Coreset~\citep{dubey2018coreset}
    & 68.00
    & \tna{}
    & 3.11\asterisk{}
    & 15.00 \\
\rowcolor{Gray}
Focused compression (5 bits, sparse)
    & 68.36
    & 88.45
    & \textbf{2.86}
    & \textbf{16.33} \\
\midrule
\thead{\resnetl}
    & \thead{Top-1} & \thead{Top-5}
    & \thead{Size (MB)} & \thead{CR (\( \times \))} \\
\midrule
INQ (5 bits)~\citep{zhou2017incremental}
    & 74.81
    & 92.45
    & 14.64\asterisk{}
    & 6.40\asterisk{} \\
ADMM (3 bits)~\citep{leng2018extremely}
    & 74.0\hphantom{0}
    & 91.6\hphantom{0}
    & 8.78\asterisk{}
    & 10.67\asterisk{} \\
ThiNet~\citep{hao2017thinet}
    & 72.04
    & 90.67
    & 16.94
    & 5.53\asterisk{} \\
Clip-Q~\citep{tung2018clipq}
    & 73.70
    & \tna{}
    & 6.70
    & 14.00\asterisk{}  \\
Coreset~\citep{dubey2018coreset}
    & 74.00
    & \tna{}
    & 5.93\asterisk{}
    & 15.80  \\
\rowcolor{Gray}
Focused compression (5 bits, sparse)
    & \textbf{74.86}
    & \textbf{92.59}
    & \textbf{5.19}
    & \textbf{18.08} \\
\bottomrule
\end{tabular}
}
\end{table}

\subsection{Computation Reduction}\label{sec:eval:compute}

Quantizing weights using FC
can significantly reduce computation complexities
in models.
By further quantizing activations
and BN parameters to integers,
the expensive floating-point multiplications
and additions in convolutions
can be replaced
with simple bit-shift operations
and integer additions.
This can be realized with
much faster software or hardware implementations,
which directly translates
to energy saving and much lower latencies
in low-power devices.
In \Cref{tab:quantize},
we evaluate the impact on accuracies
by progressively applying FQ on weights,
and integer quantizations on activations
and batch normalization (BN) parameters.
It is notable that
the final fully quantized model
achieve similar accuracies to LQ-Net.

\begin{figure}[ht]
    \centering%
    \includegraphics[width=0.8\linewidth]{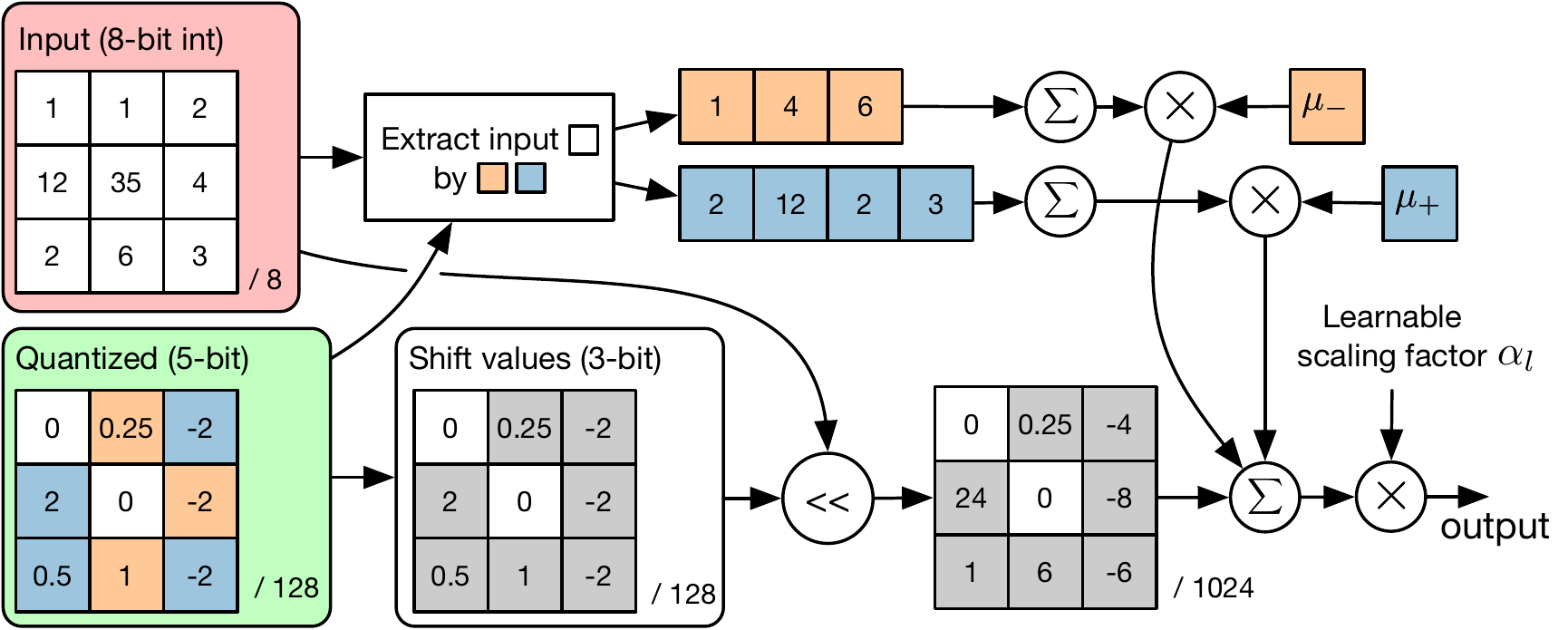}
    \caption{%
        An implementation
        of the dot-product used in convolution
        between an integer input
        and a filter quantized
        by recentralized quantization.
        The notation \( / N \)
        means the filter values
        share a common denominator \( N \).
    }\label{fig:arithmetic}
\end{figure}

\Cref{fig:arithmetic}
shows an accelerator design
of the dot-products used in the convolutional layers
with recentralized quantization for inference.
Using this, in \Cref{tab:bitops}
we provide the logic usage required
by the implementation
to compute a convolution layer
with \( 3 \times 3 \) filters with a padding size of 1,
which takes as input a \( 8 \times 8 \times 100 \) activation
and produce a \( 8 \times 8 \times 100 \) tensor output.
Additionally, we compare FQ to shift quantization,
ABC-Net~\citep{lin2017towards} and LQ-Net~\citep{zhanglq2018}.
The \#Gates indicates
the lower bound on the number
of two-input logic gates required
to implement the custom hardware accelerators
for the convolution,
assuming an unrolled architecture
and the same throughput.
Internally, a 5-bit FQ-based inference
uses 3-bit unsigned shift quantized weights,
with a minimal overhead for the added logic.
Scaling constants
\( \sigma_- \) and \( \sigma_+ \) are equal
and thus can be fused into \( \alpha_l \).
Perhaps most surprisingly,
a 5-bit FQ has more quantization levels
yet uses fewer logic gates,
when compared to ABC-Net and LQ-Net
implementing the same convolution
but with different quantizations.
Both ABC-Net and LQ-Net
quantize each weight to \( N \) binary values,
and compute \( N \) parallel
binary convolutions for each binary weight.
The \( N \) outputs are then accumulated
for each pixel in the output feature map.
In \Cref{tab:bitops},
they use \( N = 5 \) and \( 2 \) respectively.
Even with the optimal compute pattern
proposed by the two methods,
there are at least \( O(MN) \)
additional high-precision multiply-adds,
where \( M \) is the number
of parallel binary convolutions,
and \( N \) is the number of output pixels.
This overhead is much more significant
when compared to other parts of compute
in the convolution.
As shown in \Cref{tab:bitops},
both have higher logic usage
because of the substantial amount
of high-precision multiply-adds.
In contrast,
FQ has only one final learnable
layer-wise scaling multiplication
that can be further optimized out
as it can be fused into BN for inference.
Despite having more quantization levels
and a higher CR,
and being more efficient in hardware resources,
the fully quantized \resnetxviii{}
in \Cref{tab:quantize}
can still match the accuracy
of a LQ-Net \resnetxviii{}.\footnote{%
    It is also notable
    that LQ-Net used ``pre-activation'' \resnetxviii{}
    which has a 1.4\% advantage in baseline accuracy compared to ours.}

\begin{table}
\centering
\caption{%
    Comparison of the original \resnetxviii{}
    with successive quantizations
    applied on weights, activations and BN parameters.
    Each row denotes added quantization on new components.
}\label{tab:quantize}
\adjustbox{scale=1.0}{%
\begin{tabular}{lrlrl}
    \toprule
    \thead{Quantized}
        & \thead{Top-1} & \thead{\( \Delta \)}
        & \thead{Top-5} & \thead{\( \Delta \)} \\
    \midrule
    Baseline
        & 68.94 & \tna{} & 88.67 & \tna{} \\
    + Weights (5-bit FQ)
        & 68.36 & -0.58 & 88.45 & -0.22 \\
    + Activations (8-bit integer)
        & 67.89 & -1.05 & 88.08 & -0.59 \\
    + BN (16-bit integer)
        & 67.95 & -0.99 & 88.06 & -0.61 \\
    \bottomrule
\end{tabular}}
\end{table}
\begin{table}
\centering
\caption{%
    Computation resource estimates
    of custom accelerators for inference
    assuming the same compute throughput.
}\label{tab:bitops}
\adjustbox{scale=1.0}{%
\begin{tabular}{lrr}
    \toprule
    \thead{Configuration} & \thead{\#Gates} & \thead{Ratio} \\
    \midrule
    ABC-Net (5 bases, or 5 bits) & 806.1 M & \( 2.93\times \) \\
    LQ-Net (2 bits) & 314.4 M & \( 1.14\times \) \\
    Shift quantization (3 bits, unsigned) & 275.2 M & \( 1.00\times \) \\
    FQ (5 bits) & 275.6 M & \( 1.00\times \) \\
    \rowcolor{Gray}
    FQ (5 bits) + Huffman & 276.4 M & \( 1.00\times \) \\
    \bottomrule
\end{tabular}}
\end{table}

\begin{figure}[ht]
    \centering
    \includegraphics[width=0.65\linewidth]{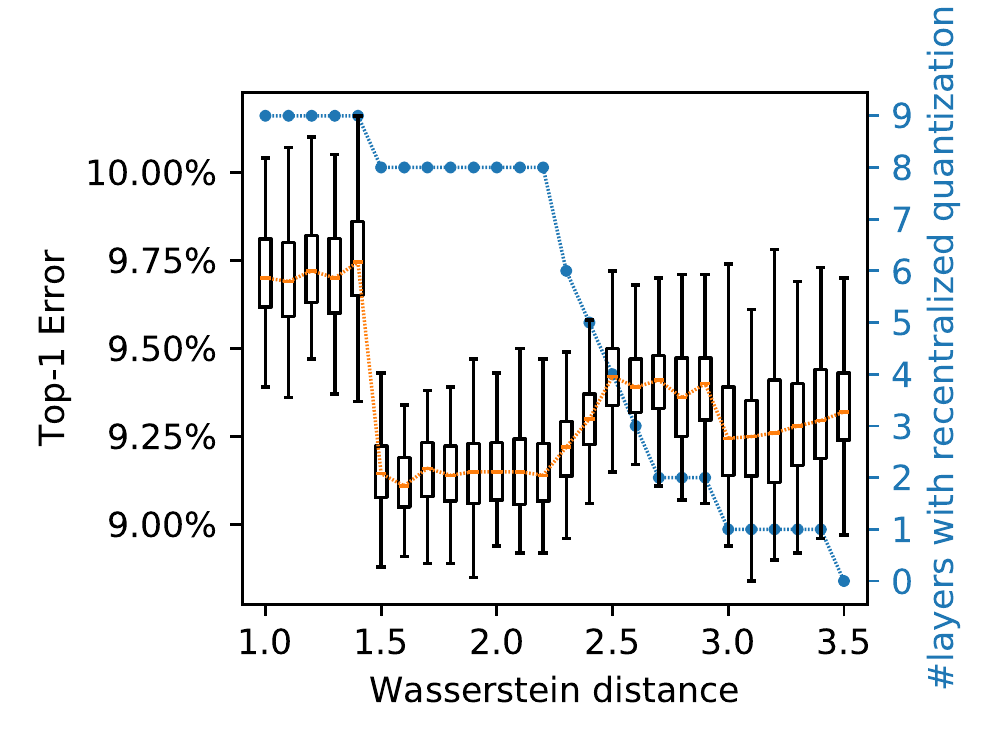}
    \caption{%
        The effect of different threshold values
        on the Wasserstein distance.
        The larger the threshold,
        the fewer the number of layers
        using recentralized quantization
        instead of shift quantization.
    }\label{fig:wdist_sweep}
\end{figure}


\subsection{Exploring the Wasserstein Separation}\label{sec:eval:wsep}

In \Cref{sec:method:wsep},
we mentioned that
some of the layers in a sparse model
may not have multiple high-probability regions.
For this reason, we use the Wasserstein distance
\( \wasserstein(c_1, c_2) \)
between the two components
in the Gaussian mixture model
as a metric to differentiate
whether recentralized or shift quantization
should be used.
In our experiments,
we specified a threshold
\( w_\mathrm{sep} = 2.0 \)
such that for each layer, if
\( \wasserstein(c_1, c_2) \geq w_\mathrm{sep} \)
then recentralized quantization is used,
otherwise shift quantization is employed instead.
\Cref{fig:wdist_sweep}
shows the impact of choosing different
\( w_\mathrm{sep} \) ranging
from 1.0 to 3.5 at 0.1 increments
on the Top-1 accuracy.
This model is a fast
\cifarx~\citep{krizhevsky2014cifar}~classifier
with only 9 convolutional layers,
so that it is possible to repeat
training 100 times for each \( w_\mathrm{sep} \) value
to produce high-confidence results.
Note that the average validation accuracy
is minimized when
the layer with only one high-probability region
uses shift quantization
and the remaining 8
use recentralized quantization,
which verifies our intuition.

\section{Conclusion}\label{sec:conclusion}

In this paper,
we exploit the statistical properties of sparse CNNs
and propose focused quantization
to efficiently and effectively
quantize model weights.
The quantization strategy uses
Gaussian mixture models to
locate high-probability regions
in the weight distributions
and quantize them in fine levels.
Coupled with pruning and encoding,
we build a complete compression pipeline
and demonstrate high compression ratios
on a range of CNNs.
In \resnetxviii,
we achieve \( 18.08 \times \) CR
with minimal loss in accuracies.
We additionally show
FQ allows a design
that is more efficient in hardware resources.
Furthermore,
the proposed quantization
uses only powers-of-2 values
and thus provides an efficient compute pattern.
The significant reductions in model sizes
and compute complexities
can translate to direct savings
in power efficiencies
for future CNN accelerators on loT devices.
Finally,
FQ and the optimized models
are fully open-source
and released to the public%
\footnote{%
    Available at:
    \href{https://github.com/deep-fry/mayo}{%
        \texttt{https://github.com/deep-fry/mayo}}.
}.

\section*{Acknowledgments}

This work is supported in part
by the National Key R\&D Program of China
(No. 2018YFB1004804),
the National Natural Science Foundation of China
(No. 61806192).
We thank EPSRC for providing
Yiren Zhao his doctoral scholarship.

\bibliographystyle{plain}
\bibliography{references}
\end{document}